\documentclass[letterpaper, 10 pt, conference]{ieeeconf}  

\IEEEoverridecommandlockouts                              

\usepackage{cite}
\usepackage{amsmath,amssymb,amsfonts}
\usepackage{algorithmic}
\usepackage{graphicx}
\usepackage{color}
\usepackage{textcomp}
\usepackage{fourier} 
\usepackage{array}
\usepackage{makecell}
\usepackage{balance}

\title{Generate What You Can't See - a View-dependent Image Generation}

\author{Karol Piaskowski$^{1}$, Rafal Staszak$^{1}$, Dominik Belter$^{1}$
\thanks{*This work was supported by the National Centre for Research and Development (NCBR) through project LIDER/33/0176/L-8/16/NCBR/2017. We gratefully acknowledge the support of NVIDIA Corporation with the donation of the Titan Xp GPU used for this research.}
\thanks{$^{1}$Institute of Control, Robotics and Information Engineering, Poznan University of Technology, Poznan, Poland
        {\tt\small dominik.belter@put.poznan.pl}}%
}

\begin{document}

\maketitle
\thispagestyle{empty}
\pagestyle{empty}

\begin{abstract}
In order to operate autonomously, a robot should explore the environment and build a model of each of the surrounding objects. A common approach is to carefully scan the whole workspace. This is time-consuming. It is also often impossible to reach all the viewpoints required to acquire full knowledge about the environment. Humans can perform shape completion of occluded objects by relying on past experience. Therefore, we propose a method that generates images of an object from various viewpoints using a single input RGB image. A deep neural network is trained to imagine the object appearance from many viewpoints. We present the whole pipeline, which takes a single RGB image as input and returns a sequence of RGB and depth images of the object. The method utilizes a CNN-based object detector to extract the object from the natural scene. Then, the proposed network generates a set of RGB and depth images. We show the results both on a synthetic dataset and on real images.
\end{abstract}

\section{Introduction}

Mobile manipulating robots, operating autonomously in human environments, have limited sensory abilities. Manipulation planning benefits from possession of a complete object model~\cite{Kopicki2015,Hillenbrand2012}. But, if the object is unfamiliar, so as to construct a complete model, the robot must either circumnavigate the object or use a wrist camera to scan it from multiple viewpoints. In contrast, humans use strong priors to complete objects by imagining occluded parts of the object from just a single view.
The reconstruction task based on sparse image data is addressed here.

\begin{figure}[t]
 \centering
\includegraphics[width=0.75\columnwidth]{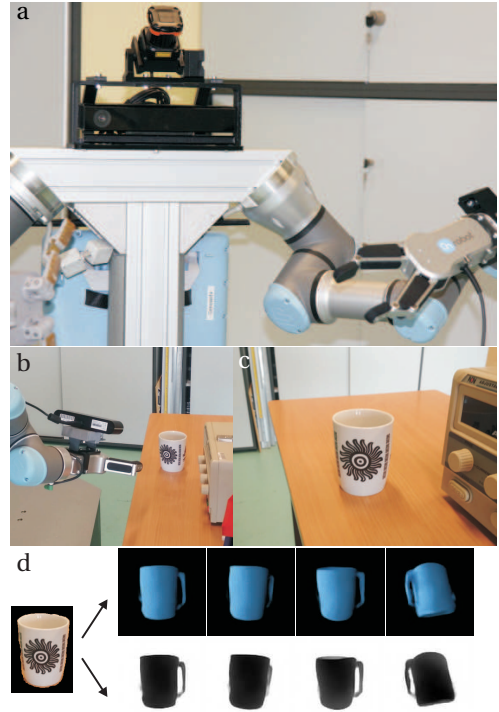}
\put(-182,267){\color{black}{a}} 
\put(-182,134){b} \put(-98,135){\color{white}{c}}
\put(-182,58){d}
\caption{Mobile manipulating robot (a) collecting information about the environment (b). Collecting images of the object from varius viewpoints is difficult due to collisions of the robot's arm with other objects -- the robot has to ``imagine'' the object from various perspectives to recover information about the object (c). Our system uses a single image to generate a set of RGB and depth images which allows to recover information about occluded parts of the object (d). Generated images can be used to find the handle of the mug and the pose of the camera where the handle is visible.}
 \label{robot40ps}
\end{figure}

In this paper, we present a system designed for use by a mobile manipulation robot. The goal is to generate a set of images of an object from desired viewpoints using a single input image. For example, in the situation presented in Fig.~\ref{robot40ps} the robot collects information about an object using the RGB-D sensor. Some parts of the object (handle, the rear surface of the mug) are occluded. To avoid careful scanning, we propose a method which allows recovering information about the object from a single view. If a grasping~\cite{Kopicki2015} or motion planning algorithm~\cite{Frank2011} algorithm needs specific information about the object's visual representation, the recovered images can be used to provide such data.

Recently, reconstruction abilities from a single view have been achieved using deep neural networks. Examples include the 3D Recurrent Reconstruction Neural Network (3D-R2N2)~\cite{Choy2016} and the 3D Generative Adversarial Network (3D-GAN)~\cite{Wu2016}. In each case, the neural network produces a 3D occupancy grid. However, fine details of the reconstructed objects are difficult to obtain directly. It is caused by the rapid growth of the computational cost and memory demands as the number of voxels increases, driven in turn by a decrease in the size of each voxel. For example, the object resolution set to $256\times256\times256$ (which is not high enough to show small details of the object) results in overall greater number of parameters and almost 17 million voxels. In contrast, neural networks are proven to perform efficiently in the image space. Thus, an image-based reconstruction has the potential to provide a better spatial resolution of a reconstructed object. The typical problems solved by the neural network are image classification and object detection~\cite{Redmon2017}, but recently they have been proven to be efficient in image synthesis~\cite{Reed2016} and scene rendering~\cite{AliEslami2018}. Thus, in this research, we propose an image-based, view-dependent approach to gather information about the object from a single view.

\begin{figure}[t]
 \centering
\includegraphics[width=0.9\columnwidth]{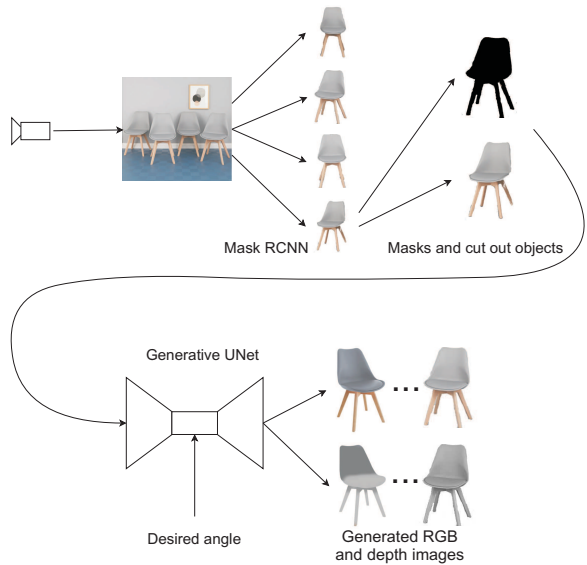}
\caption{Architecture of the single-view object reconstruction system}
 \label{pipeline}
\end{figure}

\subsection{Related Work}

Single-view images can be used for effective planning of grasping points for vacuum-based end effectors because only a single visible point of contact of suitable surface geometry is required~\cite{Mahler2018}. Along with a greater number of fingers in a gripper, the estimation of grasping points becomes more difficult. A wide variety of grasp planning methods are available. For example, Kopicki {\em et al.} \cite{Kopicki2015} presented a method for computing grasp contact points for a multi-finger robot given a partial 3D point cloud model. The grasp success rate decreases when this model is obtained from a single view. The proposed method for images generation can provide missing data and improve the grasping success rate. Another solution is to recover the 3D model and then apply grasp planning. Given a full 3D model a grasp can also be transferred to another novel object via contact warping~\cite{Hillenbrand2012}.

It is possible to recover the pose and shape of a known object from a single view using a Convolutional Neural Network (CNN), applied to the single-shot object pose estimation problem~\cite{Xiang2018}. However, most methods for object reconstruction focus on end-to-end learning a 3D voxel model of the object from a single image. A general approach, which enables the completion of a 3D shape from a single-view 3D point cloud using a CNN, was proposed by Varley et al.~\cite{Varley2017}. The network generates a 3D voxel occupancy grid from a partial point cloud and can also generalize to novel objects. The detailed mesh of the object is obtained by further post-processing of both the input point cloud and a 3D occupancy grid~\cite{Varley2017}. A similar approach to object reconstruction, based on the 3D Recurrent Reconstruction Neural Network architecture, is proposed by Choy et al.~\cite{Choy2016}. In this case, the 3D occupancy grid is obtained from an RGB image. Another approach to 3D object reconstruction is based on a set of algorithms for object detection, segmentation, and pose estimation, which fit a deformable 3D shape to the image to produce the 3D reconstruction of the object~\cite{Kar2015}.

Many objects met in manipulation tasks are symmetric. The complete shape of a partially observed object can be recovered by finding the symmetry planes and taking the scene context into account~\cite{Schiebener2016}. A similar approach to object shape prediction, based on the symmetry plane, is proposed by Bohg et al.~\cite{Bohg2011}. In contrast, a CNN-based neural network is used to complete partial 3D shapes~\cite{Dai2017}. The network operates on the 3D map of voxels and generates a high-resolution voxel grid. 

Recently, CNN has been proven to be effective in the task of rendering a whole 3D scene from few images~\cite{AliEslami2018}, image synthesis from the text~\cite{Reed2016}, or semantic image synthesis~\cite{Chen2017}, new-view image synthesis from sets of real-world, natural imagery~\cite{Flynn2016}, or image completion~\cite{Oord2016}. However, we are first to show that the sequence of 2D images of the object from a given set of viewpoints can be generated from a single image only using CNN. 

\subsection{Approach and Contribution}

\begin{figure*}[t]
\centering
  \includegraphics[width=0.99\linewidth]{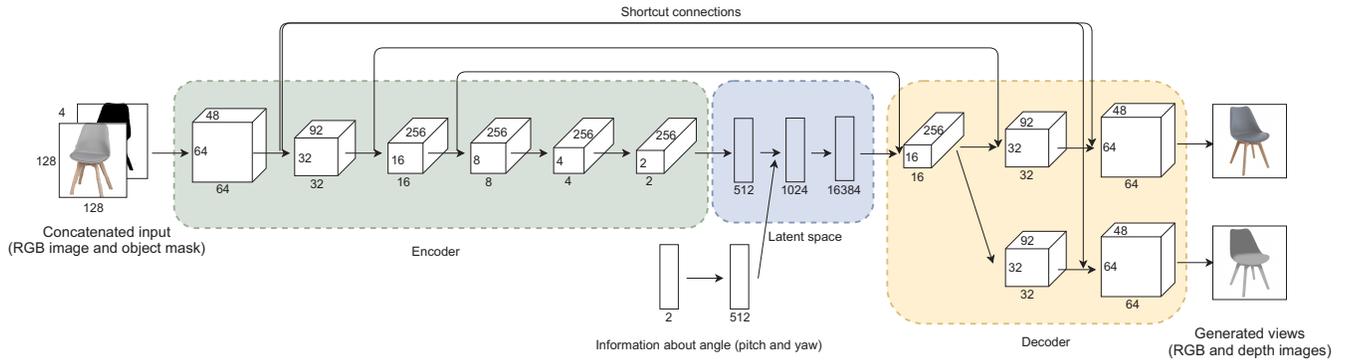}
  \caption{CNN architecture for the generation of views}
  \label{unet}
\end{figure*}

In this paper, we use 2D view-dependent approach to generate images of the object from various viewpoints. As a result, the robot can ``hallucinate'' the shape of the currently observed object (RGB and depth images) from different viewpoints.

The dominant part of our object reconstruction pipeline is based only on view dependent representations (images). This emphasis places our method in contrast to others~\cite{Choy2016, Wu2016}. Firstly, we justify our approach by the fact that human visual cortex allows performing the addressed task fairly easily. The human vision pipeline starts with position and scale-dependent representations~\cite{Spehr2015}. Then, higher layers of the perception system build 3D view-invariant models~\cite{Spehr2015}. Second, new methods from computer vision allow reconstruction of a 3D shape from the silhouette of that shape~\cite{Soltani2017}. This means that the generated 2D views can be used to generate precise point clouds of the object or 3D voxels map. Generated images can be also used to localize the relative motion of the camera by comparing the generated images from the reference view with the current camera images. It also allows finding the parts of the object which are occluded from the current view (e.g. mug handle) or to predict views during planning the motion of the robot.

\section{View-dependent Image Generation}

\begin{figure*}[t]%
  \includegraphics[width=0.99\linewidth]{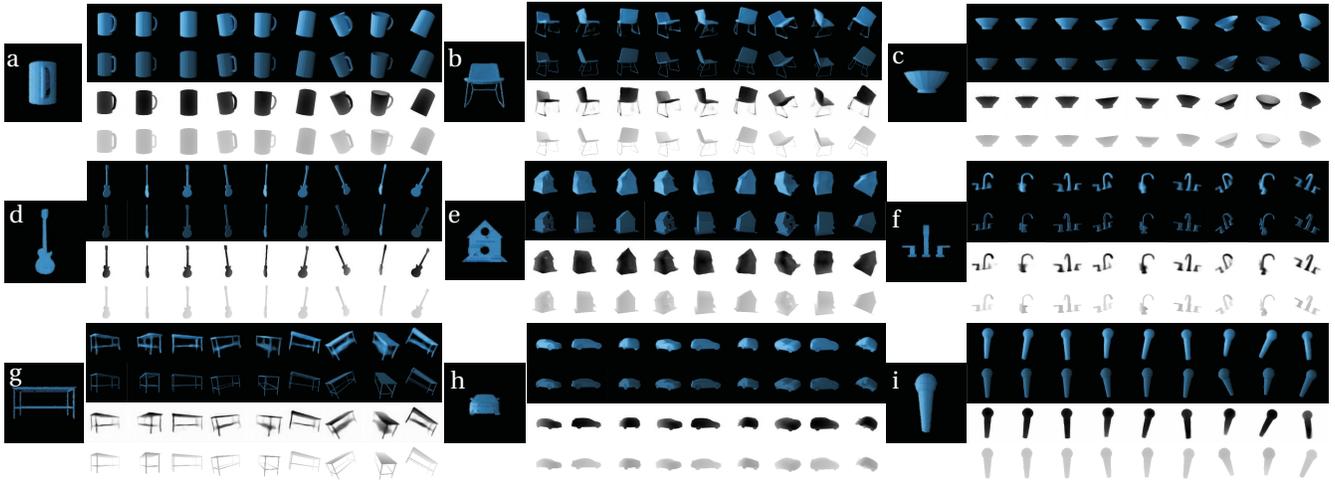}
  \put(-500,158){\color{white}{a}} \put(-333,158){\color{white}{b}} \put(-165,159){\color{white}{c}}
  \put(-499,99){\color{white}{d}} \put(-333,99){\color{white}{e}} \put(-165,97){\color{white}{f}}
  \put(-499,39){\color{white}{g}} \put(-332,36){\color{white}{h}} \put(-165,36){\color{white}{i}}
  \caption{Example images generated for the objects from the ShapeNet testing dataset. RGB images have black background while depth images have white background. Top rows in each subfigure show images generated by the neural network while the row below shows the reference images.}
    \label{generatedObjects}%
\end{figure*}

In order to generate different views of a given object, we propose a whole processing pipeline. The block diagram of the proposed method is presented in Fig.~\ref{pipeline}. The main blocks of our proposed architecture constitute two modules: object extractor and view generator.

\subsection{Object extractor}
The object extractor utilizes data from the RGB-D camera mounted on the robot. The camera provides raw information about the environment (RGB frames). We use the Mask R-CNN method~\cite{He2017} to find the 2D mask and bounding boxes of objects. After detection, the objects are cut out from the image. Our generative network operates on square images of fixed size of 
$128 \times 128$px, therefore we need to process the data obtained from the Mask R-CNN. We first scale each image in such a way that the longer side matches the required $128$px (ratio-preserving scaling). Then, in order to get a square image, we pad the images and fill them with constant background (black) color.

The same procedure is applied to the masks of detected objects. After obtaining both RGB and mask images of the object, we concatenate them and feed to the appropriate generative network. Apart from detecting objects, Mask R-CNN predicts their class labels. We utilize this information to decide which generative network should be used in the further part of the processing pipeline.

\subsection{Generative network}

Our generative network is based on the U-Net architecture ~\cite{RonnebergerFB15} (Fig.~\ref{unet}). It takes a concatenated RGB and the depth image of an object as input forwards it through the encoder and computes the latent representation of the input. The encoder is fully convolutional. It uses strides of 1 and kernels of shape 3$\times$3. Independent from the extracted features, information about the desired view angle is fed to the network. The latent representation of input and the information about angle are concatenated and forwarded through a set of fully connected layers. Before being passed to the decoder, the features are reshaped in order to match the required 3D shape for the convolutional layers. The decoder consists of a sequence of bilinear upsampling followed by standard convolutional layers. After the first convolutional layer, the network branches out into an RGB branch and a depth branch. Both branches contain two convolutional layers and are responsible for the generation of an RGB image and a depth map of the input object observed from the desired angle, respectively. 

The network utilizes shortcut connections, proposed in~\cite{RonnebergerFB15}. The motivation behind this approach is the ability of the network with a shortcut connection to keep the most important features in the latent space. Due to the limited size of the latent space, the information about less important features or small texture patches from the feature maps is stored in the encoder. These snippets of information are weighted in shortcut connections. To avoid overfitting, we rely on the concept of batch normalization~\cite{Sergey2015} and gradient clipping in the range $[-1, 1]$. Due to the abundance of data in the generated dataset, we do not use other regularization techniques and data augmentation. The weights in the network were initialized with the usage of Xavier initialization~\cite{Xavier2010}.

During experiments, we found out that a single generative network which generates images for multiple objects is difficult to obtain. Thus, we decide to use a set of small networks each dedicated to a single object class. These networks can be trained faster and with limited resources. During inference, we rely on the class predictions from Mask R-CNN in order to choose the appropriate generative network.

\subsection{Dataset}

\begin{figure}[t]%
  \includegraphics[width=0.99\linewidth]{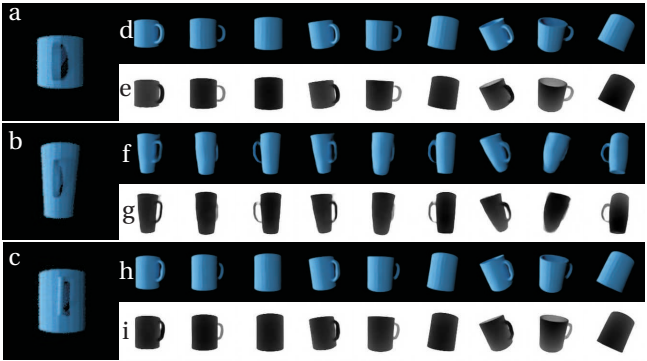}
  \put(-241,130){\color{white}{a}} \put(-241,81){\color{white}{b}} \put(-241,37){\color{white}{c}}
  \put(-199,123){\color{white}{d}} \put(-199,101){e} \put(-198,76){\color{white}{f}}
  \put(-198,55){g} \put(-199,32){\color{white}{h}} \put(-198,9){i}
  \caption{RGB and depth images generated for three different instances of the mug from the testing dataset. The objects and the orientations of the camera are not used for training. Input images are prented in subfigures (a,b,c). Generated RGB images are presented in subfigures (d,f,h), and corresponding depth images are presented in subfigures (e,g,i). Each column correspond to the same relative rotation.}
    \label{generateViews}%
\end{figure}

To the best of our knowledge, there is no large dataset of images of real objects acquired from various viewpoints. This is not surprising, taking into account an effort needed for creating such dataset. Each object should be photographed from many viewpoints in a controlled environment, with adjustable distance between camera and object, the sampling of view angles, lighting conditions and background variations. It should be noted, that some attempts to create small versions of such datasets have been made. In~\cite{Lai2011} the authors put 300 different real-world objects (belonging to 51 classes) on a turntable and photographed them with a step of about 5 degrees. A similar dataset was proposed in~\cite{Singh2014}, where 125 objects were photographed (with 600 samples per object). In~\cite{Moreels2005}, the authors collected images of 100 objects under three different lighting conditions, sampling each object 144 times. 

Unfortunately, all of the aforementioned datasets have one disadvantage: the number of different instances of objects belonging to the same class is relatively small, usually below 10. For a typical neural network, such a small amount of data is usually not sufficient. Therefore, a number of attempts have been made to utilize synthetic data, which is much easier to collect. For example, in~\cite{Aubry2014}, a method of object category detection is utilized as a solution of 2D to 3D alignment problem. The authors employ a large dataset of 3D models of artificially synthesized chairs, then successfully run it on real-world images.

In this work, we also decided to train our models on synthetic data due to its abundance. We utilized the ShapeNet dataset~\cite{Chang2015}. It contains 55 common object classes with about 51,000 unique 3D models. The objects are categorized using WordNet~\cite{Fellbaum1998} synsets, which means that each object will typically belong to several categories arranged in a hierarchy, from coarse (\textit{animal}) to fine (\textit{Siberian Husky}). The authors of ShapeNet normalized the initial position of each object. We extracted models of objects belonging to multiple categories: \textit{birdhouse, bottle, bowl, can, car, chair, faucet, guitar, lamp, microphone, mug, table}. On average, each class contains about 300 different models. Then, for each class, we rendered at most first 300 models (due to the high computational cost of rendering multiple images) at different angles. We generated both RGB images and depth maps. We sampled the pitch angle from the range 0${}^\circ$ to 30${}^\circ$ with 10${}^\circ$step and from -360${}^\circ$ to 360${}^\circ$ with 12${}^\circ$ step for the yaw angle. We did not modify the roll angle.

\subsection{Training the network}

\begin{figure}[t]%
  \includegraphics[width=0.99\linewidth]{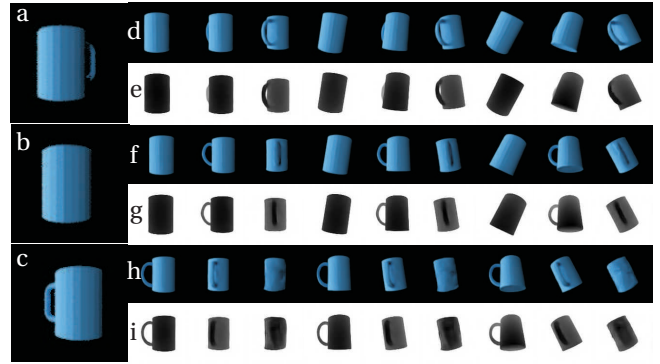}
  \put(-241,130){\color{white}{a}} \put(-241,81){\color{white}{b}} \put(-241,37){\color{white}{c}}
  \put(-199,123){\color{white}{d}} \put(-198,101){e} \put(-198,76){\color{white}{f}}
  \put(-198,55){g} \put(-199,32){\color{white}{h}} \put(-198,9){i}
  \caption{RGB and depth images generated for three different configurations of the same object from the testing dataset (c.f. caption to Fig.~\ref{generateViews}).}
    \label{generateDifferentStart}%
\end{figure}

We trained all parts of our reconstruction pipeline independently. For object extraction, we used a pre-trained Mask R-CNN and partially fine-tuned it on our data. The network was pre-trained on the COCO dataset, containing 80 classes. Unfortunately, there is no such class as \textit{can} available in COCO. Therefore, we fine-tuned Mask R-CNN on the synthetic dataset of cans. In order to prepare a dataset for Mask R-CNN, we sampled one thousand images related to things like \textit{workshops, interiors, rooms} etc. from Google. Then we randomly chose 20 different can instances. Based on that, we embedded up to 7 random objects into the randomly chosen background image. With this method, we generated 2500 training samples of cans for Mask R-CNN. 

For the generative module, we decided to use a set of 12 independent models of identical structure. Each model (neural network) is responsible for generating images of objects belonging to a single class. Each model was trained to minimize the mean square error between the target ground truth image and the generated image (RGB and depth). We set equal weights for both the RGB and depth loss. During training, we used the Adam optimizer~\cite{Kingma2014}. We trained all of the generative models on average for about 70,000 iterations, with a learning rate of 0.0005. For all layers but the final convolutional ones, we used the ReLu activation function~\cite{Hahnloser2000}.

The very important aspect of network training is data shuffling. We pair the views within each object instance randomly at every training step. However,  we never combine input and output images from various instances. Creating random pairs of mixed instances resulted in a much worse quality of generated view and lack of instance-specific detail. Therefore the network is not able to generalize the view correctly and pay attention to individual instance features.

\begin{figure}[t]
\centering
  \includegraphics[width=0.99\linewidth]{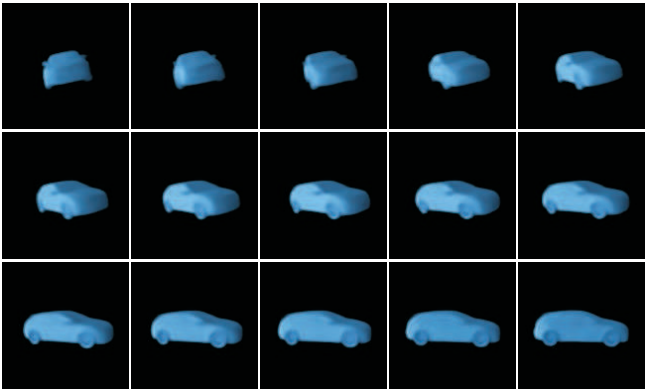}
  \caption{Colour images of a car obtained for the camera orientation changing by 6${}^\circ$. The results show that the network preserves the continuity of rotation, despite being forced to generate views of angles not available during training.}
  \label{continuity}
\end{figure} 

\section{Results} 

We are mostly interested in the visual quality of generated RGB images and depth maps. We also check the generalization capabilities of our neural network: both related to the generation of unseen objects as well as the generation of objects viewed from angles that the network was not trained on. 

The example results are presented in Fig.~\ref{generatedObjects}. In Fig.~\ref{generatedObjects} we show example images generated from the testing dataset. We tested our method on the 12 categories of objects. In Fig.~\ref{generatedObjects} we show the input image selected from the dataset and generated RGB and depth images. RGB images have a black background and depth images have a white background. The top row of the RGB and depth images are generated by the neural network, and the row below shows the reference images. Our network was not trained with various textures so the color of the object is always blue.

\begin{figure}[t]
\centering
  \includegraphics[width=0.99\linewidth]{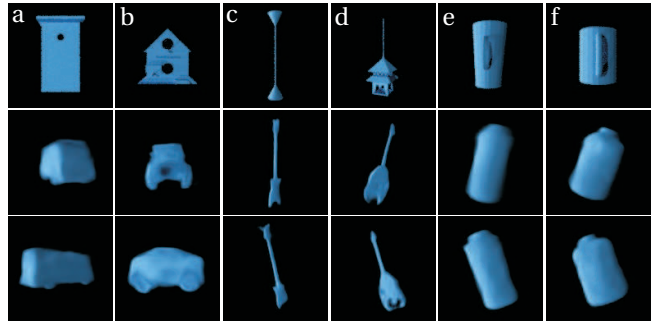}
  \put(-242,113){\color{white}{a}} \put(-201,112){\color{white}{b}} \put(-161,113){\color{white}{c}} \put(-120,112){\color{white}{d}} \put(-79,112){\color{white}{e}} \put(-38,112){\color{white}{f}}
  \caption{Results of the convertion between classes: cars generated from birdhouses (a,b), guitars generated from lamps (c,d), and bottles generated from mugs. First image in each column (a-f) shows the input image followed by images generated by the neural network.}
  \label{conversion}
\end{figure}

To test how the neural network generalizes the shape of the objects, we provided images of three different instances of mug class to the input of the neural network. Results are presented in Fig.~\ref{generateViews}. The proposed neural network can extract the visual shape of the object and generate images of this object from different perspectives despite the fact that these objects have not been shown during training the network.

\begin{table*}[t]
\caption{Average error and accuracy of the generated images}
\label{tab:error}
\begin{center}
\begin{tabular}{l|ccccccccccccc}
                        & birdhouse & bottle & bowl & can & car & chair & faucet & guitar & lamp & microphone & mug & table & average\\\hline
$e_{\rm RGB}$ [px]      & 17.38 & 11.52 & 6.76  & 12.46 & 7.19  & 14.09 & 5.61  & 4.78  & 5.38  & 8.02  & 10.40 & 11.81 & 9.62 \\
$std_{\rm RGB}$ [px]    & 4.09  & 2.66  & 1.25  & 1.48  & 2.24  & 3.53  & 1.02  & 2.16  & 1.34  & 2.03  & 1.44  & 2.76  & 2.17 \\
$acc_{\rm RGB}^\%$ [\%] & 93.19 & 95.48 & 97.35 & 95.11 & 97.18 & 94.47 & 97.80 & 98.13 & 97.89 & 96.85 & 95.92 & 95.37 & 96.2 \\
$e_{\rm d}$ [px]        & 8.10  & 3.11  & 5.10  & 5.09  & 5.53  & 7.67  & 2.21  & 0.28  & 0.77  & 1.12  & 5.90  & 7.10  & 4.33 \\
$std_{\rm d}$ [px]      & 4.64  & 2.01  & 1.20  & 1.93  & 3.27  & 2.95  & 2.34  & 0.74  & 0.61  & 1.30  & 3.58  & 2.72  & 2.21 \\
$acc_{\rm d}^\%$ [\%]   & 96.82 & 98.78 & 98.00 & 98.01 & 97.83 & 96.99 & 99.14 & 98.13 & 99.70 & 99.56 & 97.69 & 97.21 & 98.16\\
\end{tabular}
\end{center}
\end{table*}

We also verified how the neural network generates images of the same object observed from different viewpoints. In Fig.~\ref{generateDifferentStart} we show three different images of the same object and sequences of RGB and depth images from these input images. The most interesting example is presented in Fig.~\ref{generateDifferentStart}b. In the input image, the handle is not visible. However, the neural network can generate the images of the mug with the handle when we generate images from different viewpoints. The shape and size of the handle are slightly different than the real handle but the neural network can correctly predict that the handle is located on the occluded side of the mug.

\begin{figure}[t]
\centering
  \includegraphics[width=0.99\linewidth]{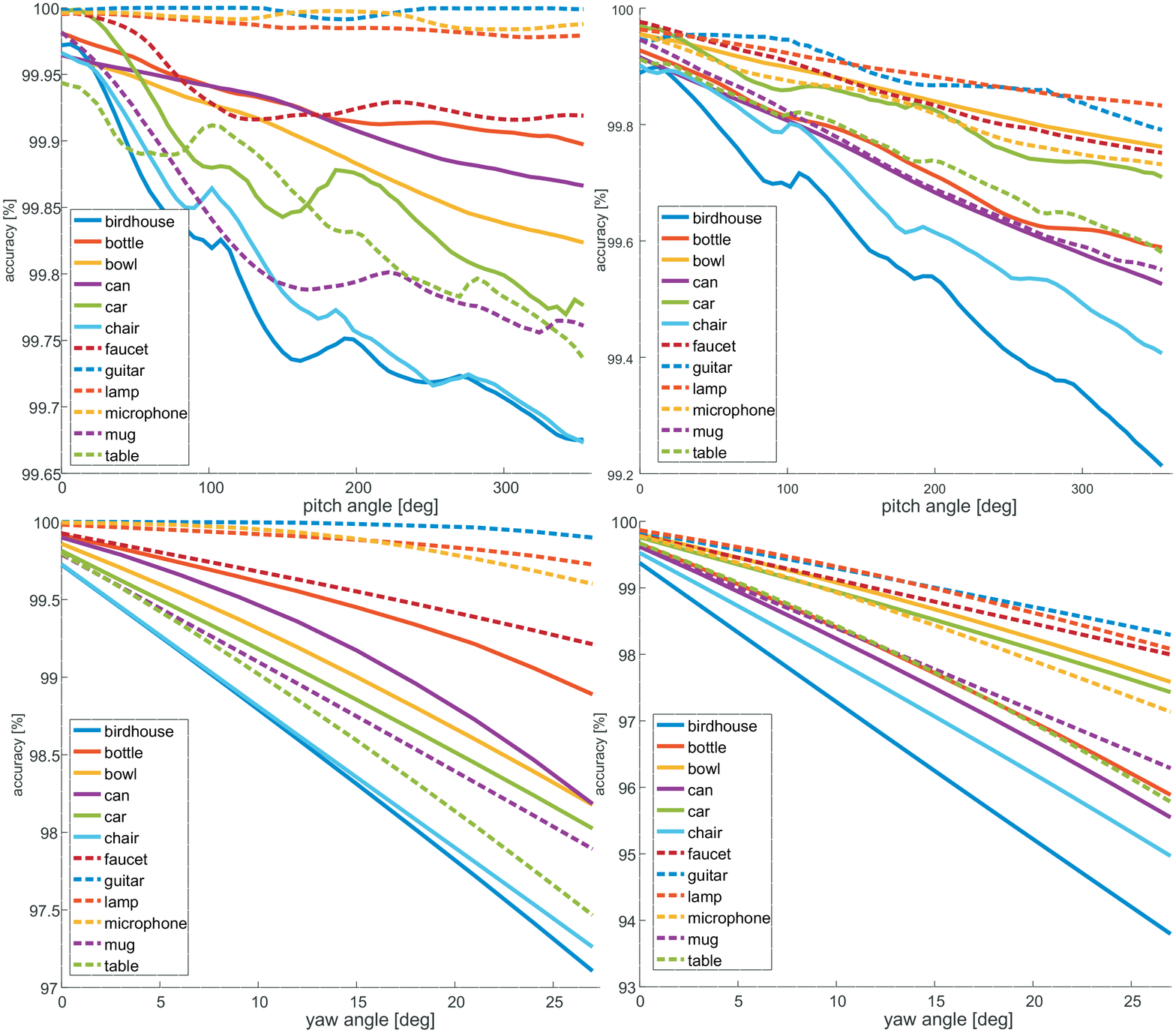}
  \put(-240,202){a} \put(-120,201){b} \put(-240,93){c} \put(-120,95){d}
  \caption{Dependency between accuracy of the generated depth (a,c) and RGB (b,d) images and the rotation angles (pitch and yaw)}
  \label{rotAccuracy}
\end{figure}

To show the properties of our method, we present how the network generates images for orientations which have never been presented to the network. The results of the experiment presented in Fig.~\ref{continuity} show that the model is able to interpolate between the training angle samples. The yaw angle for the training dataset was changed by 12${}^\circ$. The images presented in Fig.~\ref{continuity} are generated for the camera poses which differ by 6${}^\circ$. It means that the odd images are obtained for the angle presented during the training phase and the remaining images are obtained for orientation of the camera not used for the training and results are interpolated by the network. It can be clearly seen that the continuity of angle space is preserved. This fact is interesting when we take into account that the autoencoder used during training was not designed to preserve the space continuity (as opposed to, for example, variational autoencoders).

\begin{figure}[t]
\centering
  \includegraphics[width=1.0\linewidth]{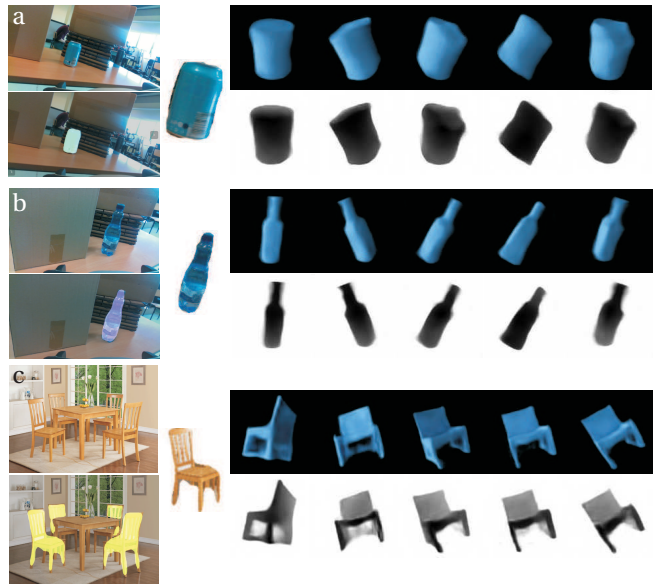}
  \put(-244,212){\color{white}{a}} \put(-244,141){\color{white}{b}} \put(-244,76){c}
  \caption{Results obtained for the real objects: a can (a), a bottle (b), and a chair (c). The first two images are obtained on the robot with an Intel RealSense D435 camera. For each object we show the input image, the image with objects detected by Mask R-CNN, the object extracted from the RGB image, the generated RGB (top row) and depth (bottom row) views}
  \label{finalResults}
\end{figure}

In the next experiment, we checked what would happen if the Mask R-CNN misclassifies the object. Thus, we provide the images of a birdhouse to the neural network related to cars (Fig.~\ref{conversion}a and Fig.~\ref{conversion}b), a lamps to the guitar model (Fig.~\ref{conversion}c and Fig.~\ref{conversion}d), and mugs to the bottle model (Fig.~\ref{conversion}e and Fig.~\ref{conversion}f). The neural network can properly generate the orientation of the object. It means that it's easier for the neural network to model the transformation of rigid objects. More difficult is the generation of the object's shape. However, it is visible how the neural network mixes the input object with the model of the object stored in the neural network producing reasonable and interesting images (Fig.~\ref{conversion}).

To provide qualitative results, we also compare the images generated by the neural network $I^{\rm gen}$ and the reference images obtained from the 3D model of the object $I^{\rm ref}$. The error between two images is computed as follows:

\begin{equation}
\label{error}
e= \frac{1}{NML} \sum_{l=1}^{L} \sum_{n=1}^{N} \sum_{m=1}^{M} |I_{l,n,m}^{\rm gen} - I_{l,n,m}^{\rm ref}|,
\end{equation}

where $N \times M$ is the size of the image, $L$ is the number of layers, $I_{n,m}^{\rm gen}$ and $I_{n,m}^{\rm ref}$ are the corresponding pixels of the reference image and the image generated by the neural network, respectively. The number of layers $L$ for the depth image is 1 and the $L$ value is set to 3 for the RGB images. We also compute the accuracy of the obtained RGB and depth images:

\begin{equation}
\label{accuracy}
acc = \frac{1-e}{255} \cdot 100\%,
\end{equation}
which is normalized by the maximal value of the image (255).

The accuracy of the proposed method is presented in Tab.~\ref{tab:error}. The number of testing objects for each category is the same. The results are obtained for the first five instances of test objects from each category of the ShapeNet dataset. For each selected object we generate five different images which differ with the observation angle. The images are generated by sampling the pitch angle from the range 0${}^\circ$ to 30${}^\circ$ with 3${}^\circ$ step and from 0${}^\circ$ to 360${}^\circ$ with 6${}^\circ$ step for the yaw angle. It means that we have 300 input images and we generate 180 000 images for comparison. Surprisingly, we obtained the best accuracy for complex objects like cars, faucets, and guitars. The objects like chairs and tables are more difficult for the proposed neural network. The biggest error is caused by the generated legs of these object. They are very often bent or vague. The lowest accuracy is obtained for the birdhouse class. This is mainly because the testing objects differ significantly from the training dataset. However, the results are still satisfactory (Fig.~\ref{generateViews}).

We also checked how the accuracy of the generated RGB and depth images depend on the rotational distance between input image viewpoint and the reference viewpoint. The results are presented in Fig.~\ref{rotAccuracy}. Unsurprisingly, the accuracy decreases when the rotation angle between the input and generated image increase but stays at a reasonable level.

Finally, we evaluated how well our model performs on real data, unavailable in the synthetic dataset. The example results can be seen in Fig.~\ref{finalResults}. These results confirm the conclusions drawn from the synthetic data. Note that our neural network was not trained with textured objects and does not generate textures on the output images.

\section{Conclusions and Future Work}

In this paper, we present a method which generates a sequence of RGB and depth images from a single RGB input image. We proposed the whole processing pipeline, which extracts the objects from the raw input image, generates a set of RGB and depth viewpoints of the query object. In the paper, we show that the proposed neural network is capable of interpolation to viewpoints not used during training and generate models of novel objects. The proposed method is especially important in the field of mobile and manipulating robots. With the proposed method the robot can better understand the spatial properties of objects, without the need for complete scanning, which is time-consuming and sometimes impossible to perform. Our method works end to end without human supervision.

The method has also some limitations which we are going to deal with in the future. The neural network can't handle the texture of the object. We also failed to train a single neural network which can generate images for different object classes. We are convinced that these problems can be solved by investigating the architecture of the proposed neural network and involvement of more computational resource. In the future, we are also going to use the generated images to reconstruct the 3D model of the object and estimate the motion of the camera by comparing generated and current camera images.

\end{document}